\documentclass[11pt]{article}

\usepackage[final]{acl}
\usepackage{times}
\usepackage{latexsym}
\usepackage[T1]{fontenc}
\usepackage[utf8]{inputenc}
\usepackage{microtype}
\usepackage{inconsolata}
\usepackage{booktabs}
\usepackage{array}
\usepackage{tabularx}
\usepackage{graphicx}
\usepackage{placeins}
\usepackage{amsmath}
\usepackage{amssymb}
\usepackage{xcolor}
\usepackage{url}
\usepackage{tikz}
\usepackage{fontawesome5}
\usepackage{pgfplots}
\pgfplotsset{compat=1.17}
\usepgfplotslibrary{groupplots}
\usetikzlibrary{matrix,positioning,arrows.meta,fit,backgrounds}

\newcommand{\method}{MechSparse}
\newcommand{\methodc}{MechSparse-C}

\newcommand{\thead}[1]{\textbf{#1}}
\newcommand{\upbetter}{\textsuperscript{\ensuremath{\uparrow}}}
\newcommand{\downbetter}{\textsuperscript{\ensuremath{\downarrow}}}
\newcolumntype{L}[1]{>{\raggedright\arraybackslash}p{#1}}
\newcolumntype{C}[1]{>{\centering\arraybackslash}p{#1}}

\title{\method{}: Mechanism-Guided Sparse PEFT Selection Is Task-Shaped\thanks{Anonymous artifacts: \url{https://anonymous.4open.science/r/MechSparse-C92D/README.md}.}}

\author{
  Son Ha Xuan\textsuperscript{1,$*$} \quad
  Phat T. Tran-Truong\textsuperscript{2,$*$} \quad
  Xuan-Bach Le\textsuperscript{2,$\dagger$} \\[3pt]
  \textsuperscript{1}RMIT University, Ho Chi Minh City, Vietnam \\
  \textsuperscript{2}Faculty of Computer Science and Engineering, Ho Chi Minh City University of Technology (HCMUT), \\
  VNU-HCM, Ho Chi Minh City, Vietnam \\[3pt]
  \texttt{ha.son@rmit.edu.vn} \quad \texttt{\{phatttt, lexuanbach\}@hcmut.edu.vn} \\[3pt]
  \textsuperscript{$*$}Equal contribution. \quad \textsuperscript{$\dagger$}Corresponding author.
}

\begin{document}
\maketitle

\begin{abstract}
Mechanistic interpretability identifies sparse subsets of heads and MLP blocks that carry specific behaviors. We ask whether such causal signals can guide where to place a small PEFT budget more effectively than the cheap heuristics practitioners already use. \method{} scores attention heads and MLP blocks by normalized activation-patching recovery on clean/corrupted probes and trains LoRA/QLoRA only on the selected sites; \methodc{} adds bounded credit for small within-layer joint subsets.

We compare against random, magnitude, activation-norm, and gradient/Fisher on Ministral-8B/NF4 in three cells: Swahili span-JSON information extraction (IE) at $b{=}0.25\%$ and $1.0\%$, and English$\to$Swahili machine translation (MT) at $b{=}1.0\%$. The causal selectors never win the primary metric. On the headline IE cell (3 seeds, paired-bootstrap CIs over $600$ predictions), \methodc{} beats random by $+0.079$ span+type F1 and gradient/Fisher by $+0.174$, but trails activation-norm by $0.028$, with the smallest cross-seed std ($\pm 0.003$). On MT all four selectors lie within $0.30$ BLEU and every paired CI includes zero. A schema-versus-span decomposition explains the IE gap: activation-norm captures the rigid JSON routine, while causal scores track content-sensitive sites. We distill a preliminary diagnostic -- prefer activation-norm when output structure dominates, treat causal selectors as a hypothesis for content-dominated tasks -- and release masks, scores, predictions, and evaluation files for direct replay.
\end{abstract}

\begin{figure*}[t]
\centering
\begin{tikzpicture}[
  every node/.style={font=\scriptsize, align=center},
  card/.style={rectangle, draw=#1!70!black, rounded corners=3pt, thick, minimum height=1.15cm, text width=2.05cm, inner sep=3pt, fill=#1!7},
  card/.default=blue,
  arrow/.style={-{Stealth[length=2mm]}, thick, draw=gray!70!black},
  group/.style={draw=#1!65!black, rounded corners=4pt, dashed, inner sep=5pt},
  group/.default=gray,
]
\node[card=blue] (probes) {\textcolor{blue!70!black}{\Large\faIcon{file-alt}}\\[-0.15em]\textbf{Probe pairs}\\[-0.15em]\scriptsize filtered span-JSON};
\node[card=purple, right=0.35cm of probes] (patch) {\textcolor{purple!70!black}{\Large\faIcon{bolt}}\\[-0.15em]\textbf{Patch}\\[-0.15em]\scriptsize normalized recovery};
\node[card=teal, right=0.35cm of patch] (circuit) {\textcolor{teal!70!black}{\Large\faIcon{project-diagram}}\\[-0.15em]\textbf{Credit}\\[-0.15em]\scriptsize joint subsets};
\node[card=orange, right=1.05cm of circuit] (mask) {\textcolor{orange!75!black}{\Large\faIcon{th}}\\[-0.15em]\textbf{Sparse mask}\\[-0.15em]\scriptsize capped budget};
\node[card=orange, below=0.62cm of mask] (peft) {\textcolor{orange!75!black}{\Large\faIcon{microchip}}\\[-0.15em]\textbf{Masked QLoRA}\\[-0.15em]\scriptsize heads/MLPs};
\node[card=green, below=0.62cm of circuit] (eval) {\textcolor{green!55!black}{\Large\faIcon{chart-line}}\\[-0.15em]\textbf{Evaluate}\\[-0.15em]\scriptsize schema + span};
\node[card=gray, left=0.35cm of eval] (claim) {\textcolor{gray!70!black}{\Large\faIcon{clipboard-check}}\\[-0.15em]\textbf{Audit}\\[-0.15em]\scriptsize eval files};
\node[card=red, left=0.35cm of claim] (report) {\textcolor{red!70!black}{\Large\faIcon{compass}}\\[-0.15em]\textbf{Decision rule}\\[-0.15em]\scriptsize task-shaped};
\node[group=blue, fit=(probes) (patch) (circuit), label={[font=\scriptsize\bfseries,text=blue!55!black]above:Mechanism scoring}] {};
\node[group=orange, fit=(mask) (peft), label={[font=\scriptsize\bfseries,text=orange!65!black]right:Sparse transfer}] {};
\node[group=green, fit=(eval) (claim) (report), label={[font=\scriptsize\bfseries,text=green!45!black]below:Evidence and policy}] {};
\draw[arrow] (probes) -- (patch);
\draw[arrow] (patch) -- (circuit);
\draw[arrow] (circuit) -- (mask);
\draw[arrow] (mask) -- (peft);
\draw[arrow] (peft) -- (eval);
\draw[arrow] (eval) -- (claim);
\draw[arrow] (claim) -- (report);
\end{tikzpicture}
\caption{The \method{} pipeline. We score each attention head and MLP block by activation patching on filtered clean/corrupt probes (\emph{Mechanism scoring}, blue), pick a sparse mask under a matched budget and train QLoRA only on the selected components (\emph{Sparse transfer}, orange), then evaluate and distill a selector decision rule (\emph{Evidence and policy}, green). Cells: Swahili span-JSON IE at $b{=}0.25\%$ and $1.0\%$, and English$\to$Swahili MT at $b{=}1.0\%$.}
\label{fig:pipeline}
\end{figure*}
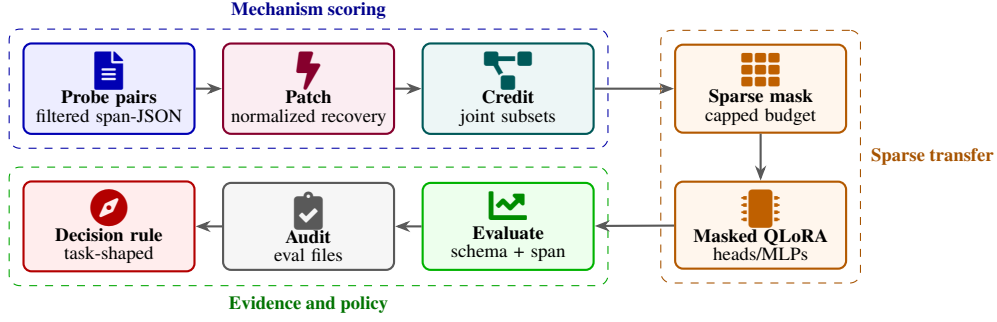

\section{Introduction}

Parameter-efficient fine-tuning (PEFT) has made multilingual adaptation cheaper, but most pipelines still decide \emph{where} to train by architectural convention: a LoRA or adapter baseline updates the same projections in every layer, regardless of task or language \citep{houlsby2019parameter,hu2022lora,dettmers2023qlora,pfeiffer2020adapterhub}. Mechanistic interpretability suggests a sharper alternative. If a task depends disproportionately on a small set of heads or feed-forward blocks \citep{voita2019analyzing,michel2019sixteen,geva2021transformer,meng2022locating,vig2020investigating,zhang2023patching}, the adapter budget should target those components rather than spread across a fixed template.

The question is not whether causal scores rank components in isolation, but whether they beat the cheap heuristics practitioners actually use -- random, weight magnitude, and activation norm. Patch scores depend on the corruption operator, the normalization denominator, and the output metric \citep{zhang2023patching,heimersheim2024patching}; rather than a flaw, this dependence reveals which task objective a selector optimizes for. Prior sparse-PEFT work has compared optimization-derived masks against random or architectural baselines \citep{zhang2023adalora,ding2023sora}; we instead ask whether a \emph{causal} signal identifies the sites that sparse fine-tuning should update.

\method{} makes the comparison directly. We build clean/corrupted probe pairs, score attention heads and MLP blocks by normalized patch recovery, optionally add bounded within-layer joint-subset credit (\methodc{}), select a sparse mask, and train only adapters attached to the selected components (Figure~\ref{fig:pipeline}). We evaluate against random, magnitude, activation-norm, and gradient/Fisher on three Ministral-8B/NF4 cells: Swahili span-JSON IE at budgets $0.25\%$ and $1.0\%$, and English$\to$Swahili MT at $1.0\%$. The headline IE cell uses three end-to-end seeds with pooled paired-bootstrap CIs over $600$ predictions.

\paragraph{Finding.}
Selector quality is \emph{task-shaped}, and our causal selectors do not win the primary metric in any of the three cells. On the headline IE cell ($b{=}0.25\%$, 3 seeds), \methodc{} sits $+0.079$ span+type F1 above random and $+0.174$ above gradient/Fisher, with both paired-bootstrap CIs strictly positive; the gap to magnitude is small and its CI crosses zero. Activation-norm leads \methodc{} by $0.028$ (CI fully negative) because this IE setting rewards rigid schema and type-string routines as much as content localization, yet \methodc{} carries the tightest cross-seed std on the primary metric ($\pm 0.003$). The pattern persists across cells: \methodc{} beats random on three of four IE metrics everywhere but trails magnitude on span+type F1 at $b{=}1.0\%$, while on MT we see no statistically reliable separation between any pair of selectors (Appendix~\ref{app:mechsparse_remaining}). A seed-29 diagnostic clears our preregistered $0.45$ Jaccard/correlation floor on all six measures (Figure~\ref{fig:stability}), so the causal ranking is reproducible even where it loses downstream.

\paragraph{Interpretation.}
The IE pilot bundles two objectives into one metric: produce content-correct entities, and emit rigid JSON with valid braces, keys, list structure, and type labels. Clean/corrupt patch scores target the first; activation-norm picks up components that fire reliably for the schema routine regardless of input (Figure~\ref{fig:schema_vs_span}). On MT, where output format is loose, the four selectors converge (Section~\ref{sec:analysis} discusses a competing floor-effect reading). We therefore offer a preliminary rule: prefer activation-norm when the task is dominated by ``always-on'' output structure, and treat causal or probe-delta selectors as a hypothesis for content-dominated tasks not yet demonstrated here.

\paragraph{Contributions.}
\begin{itemize}
  \item \method{}, a reproducible mechanism-guided sparse PEFT pipeline that scores components by normalized patching, optionally adds bounded joint-subset credit, and trains masked QLoRA only on selected sites under matched per-layer caps.
  \item A head-to-head comparison against random, magnitude, activation-norm, and gradient/Fisher under matched budgets: causal selection beats random and gradient/Fisher significantly, ties magnitude, but wins no primary metric across our three cells -- a negative-result \emph{diagnostic} rather than a method win.
  \item A schema-versus-span decomposition that explains the IE gap and yields a preliminary task-shaped selector rule, with a seed-29 stability diagnostic clearing our preregistered $0.45$ Jaccard/correlation floor; we release masks, scores, predictions, and the 3-seed evaluation files for direct replay.
\end{itemize}

\paragraph{Roadmap.}
Section~\ref{sec:method} defines the \method{} pipeline and the \methodc{} extension; Section~\ref{sec:experiments} introduces the task cells, baselines, and seed protocol; Section~\ref{sec:results} reports the headline IE pilot and cross-cell consistency; Section~\ref{sec:analysis} develops the schema-vs-span decomposition, the stability diagnostic, and the selector decision rule. Related work and the conclusion close out the body.

\section{Method}
\label{sec:method}

\paragraph{Setup.}
Let $M$ be a frozen decoder-only model whose component set $\mathcal{C}$ consists of attention heads $H_{\ell,h}$ and MLP blocks $F_\ell$. We choose a trainable subset $S\!\subseteq\!\mathcal{C}$ with a sparse PEFT selector and learn adapter parameters $\phi_S$ while $M$ stays frozen. The selector solves the central allocation problem: which internal sites should receive the limited adapter budget. We use Ministral-8B Instruct quantized with NF4 and train QLoRA adapters (exact identifier in Appendix~\ref{app:hyperparams}).

\paragraph{Probes.}
We construct clean/corrupted probe tuples $(x_i,\tilde{x}_i,y_i,\tilde{y}_i)$. For span-JSON IE, we define the behavioral contrast by swapping an entity span and/or type in the input and updating the target JSON correspondingly. The contrast asks which components move probability back toward the clean target when the input entity changes. We run all probing and patching on the quantized model so scores reflect the deployment circuit. We retain probes via a competence filter (clean prob.\ $\ge 0.55$, corrupt $\le 0.45$, $\ge 64$ probes), then apply patching with a $64$-probe limit.

\paragraph{Component scoring.}
For each component $c$ we estimate recoverability with activation patching: we replace the corrupted-run activation at $c$ with the clean-run activation and measure recovery of the clean target. A component scores high when this intervention restores the target more than the corrupted run alone. Let $q(M,x,y)=\log\Pr_M(y\!\mid\!x)$. The raw patch score is
\begin{equation}
r_c(i) = q(M_{c\leftarrow\mathrm{clean}}, \tilde{x}_i, y_i) - q(M, \tilde{x}_i, y_i).
\end{equation}
We normalize by the recoverable range and use $\epsilon{=}0.05$:
\begin{equation}
\label{eq:score}
s_c
= \mathbb{E}_{i\in\mathcal{P}_{t}}
\left[
\frac{r_c(i)}
{\big(q(M,x_i,y_i)-q(M,\tilde{x}_i,y_i)\big)+\epsilon}
\right].
\end{equation}
We bootstrap scores with $100$ samples. The same pipeline also gives us a probe-delta selector that ranks components by clean/corrupt probe sensitivity without the \methodc{} joint-subset mix.

\paragraph{Bounded joint-subset extension (\methodc{}).}
Single-component patching can underestimate distributed circuits, the hydra effect \citep{zhang2023patching}. We augment single scores with within-layer joint subsets in \methodc{}: top fraction $0.20$, subset size $\le 2$, $\le 8$ subsets per layer, mix coefficient $\lambda{=}0.4$. The suffix \emph{-C} marks this circuit-level credit, since we give bounded credit to local component sets that recover the clean behavior jointly.

\paragraph{Sparse mask and LoRA placement.}
Given a target budget, we translate scores into trainable sites by selecting the highest-scoring components subject to a per-layer cap of $40\%$ and a component-type floor of $20\%$. We set $b{=}0.25\%$ on the headline cell with head rank $4$ and MLP rank $8$ ($\sim$$15$M trainable). The base model uses grouped-query attention ($32$ query heads, $8$ KV groups), so we attach LoRA at the projection level when any head in the relevant group is selected, with GQA K/V mask deduplication to avoid double-counting parameters. MLP-selected blocks get LoRA on gate/up/down projections. We acknowledge that this placement is coarser than the per-head score (Section~\ref{sec:limitations}).

\paragraph{Training, evaluation, and decision criteria.}
We train adapters by minimizing sequence loss over span-JSON data: one epoch, lr $2\!\times\!10^{-4}$, grad.\ accum.\ $16$, max length $768$, bf16, test-time max $96$ new tokens. We report JSON validity, exact-schema match, span F1, and span+type F1 (primary). We preregister two criteria. For \emph{quality}, a mechanism-guided mask should be competitive with matched-budget heuristics and materially improve over weak signals. For \emph{stability}, single-score cross-seed Jaccard/correlations must exceed $0.45$. We use the stability criterion as a reproducibility check and interpret the quality criterion through the task-shaped decision rule in Section~\ref{sec:analysis}.

\section{Experimental Setup}
\label{sec:experiments}

\paragraph{Task cells.}
We evaluate selectors across three task cells (Table~\ref{tab:cells}). We treat Swahili span-JSON IE at budget $0.25\%$ as the headline cell, with a strict-JSON prompt (zero-based exclusive offsets) and GQA K/V mask deduplication. We add two supplementary cells, Swahili IE and English$\to$Swahili MT at budget $1.0\%$, as cross-cell checks. The supplementary IE cell uses a looser prompt, and both supplementary cells predate GQA dedup. We re-score the supplementary IE predictions with the same evaluator we use on the headline cell, and MT BLEU/chrF is unchanged across iterations. We use the headline cell for seed-replicated claims and the supplementary cells to test within-cell selector ordering (Appendix~\ref{app:mechsparse_remaining}).

\begin{table}[t]
\centering
\footnotesize
\setlength{\tabcolsep}{2.5pt}
\begin{tabularx}{\linewidth}{@{}L{1.15cm}L{1.75cm}C{0.85cm}C{0.90cm}C{1.20cm}C{0.45cm}@{}}
\toprule
\thead{Cell} & \thead{Task} & \thead{Bud.\downbetter} & \thead{Metric} & \thead{IE prompt} & \thead{N\upbetter} \\
\midrule
IE-prim. & Span-JSON IE & 0.25\% & S+T F1 & Strict & 3 \\
IE-supp. & Span-JSON IE & 1.0\% & S+T F1 & Loose & 1 \\
MT       & En$\!\to\!$Sw MT & 1.0\% & BLEU & --- & 1 \\
\bottomrule
\end{tabularx}
\caption{Three task cells. Bud.\ = LoRA budget, $N$ = end-to-end seeds, S+T F1 = span+type F1. Strict prompt = zero-based exclusive offsets and GQA K/V mask dedup. Loose prompt = supplementary-cell prompt without GQA dedup.}
\label{tab:cells}
\end{table}

\paragraph{Data.}
For IE, we convert each MasakhaNER \texttt{swa} example \citep{adelani2022masakhaner} to a span-JSON instance asking the model to emit a JSON object with an \texttt{entities} list (\texttt{span}, \texttt{type}, \texttt{start}, \texttt{end}). This format lets us measure content recovery and schema preservation in the same output. The headline IE cell uses $1024$ training, $200$ validation, and $200$ test examples. For MT we use a FLORES-style English$\to$Swahili split with BLEU/chrF eval. Per-run dataset manifests are released with the artifacts.

\paragraph{Model and adapter budget.}
We use Ministral-8B Instruct with NF4 quantization and bf16 compute across all cells ($36$ transformer layers, $32$ attention heads, $8$ key-value heads, hidden size $4096$, intermediate size $12288$, with the exact identifier in Appendix~\ref{app:hyperparams}). The headline cell uses head LoRA rank $4$ and MLP LoRA rank $8$ ($\sim$15M trainable). The supplementary cells use the same LoRA shape at a $1.0\%$ budget ($\sim$60M trainable).

\paragraph{Compared selectors.}
\begin{itemize}
  \item \textbf{Random}: components sampled uniformly under the matched budget.
  \item \textbf{Magnitude}: components ranked by weight magnitude.
  \item \textbf{Activation-norm}: mean activation magnitude over $64$ task prompts from the same train-distribution split used by the probe statistics, with no test data.
  \item \textbf{\methodc{}}: normalized clean/corrupt patch recovery with a bounded within-layer joint-subset mix (Section~\ref{sec:method}).
  \item \textbf{Probe-delta}: clean/corrupt probe sensitivity without the joint mix (headline cell).
  \item \textbf{Gradient/Fisher}: diagonal Fisher-information score of adapted parameters on $64$ task prompts under matched NF4/QLoRA (headline cell).
  \item \textbf{Hybrid/gated}: exploratory combinations of activation and causal scores (Appendix~\ref{app:mechsparse_remaining}).
\end{itemize}
AdaLoRA-style adaptive-rank and Taylor-importance are complementary rank-allocation baselines for the next comparison (Section~\ref{sec:limitations}).

\paragraph{Metrics and protocol.}
We evaluate IE with JSON validity, exact-schema match, span F1, and span+type F1 (primary). MT uses BLEU and chrF \citep{popovic2015chrf,post2018sacrebleu}. For the headline IE cell we run three end-to-end training seeds (13, 29, 47) for each of random, magnitude, activation-norm, \methodc{}, and gradient/Fisher, reporting per-metric mean$\pm$std plus paired bootstrap CIs over the $600$ pooled predictions. A single-score \method{} diagnostic on seed 29 vs.\ base checks score reproducibility (Figure~\ref{fig:stability}). We release per-method evaluation files, a 3-seed aggregate, masks, selector scores, and recovered MT prediction bundles.

\paragraph{Selector cost.}
On the L40S GPU we use for the headline run, magnitude takes $\sim$$11$s (a weight scan), activation-norm $\sim$$19$s ($64$ forward passes), and gradient/Fisher a short training pass. Patch-based scoring is the expensive step at $\sim$$78$min single-component + $\sim$$22$min joint-subset $=$ $\sim$$1.7$ GPU-hours per cell. Per-selector training plus evaluation runs in $\sim$$10$--$15$min. We amortize this cost across budgets and PEFT seeds whenever we reuse the mask, and we turn the cost profile into a deployment rule in Section~\ref{sec:analysis}.

\section{Results}
\label{sec:results}

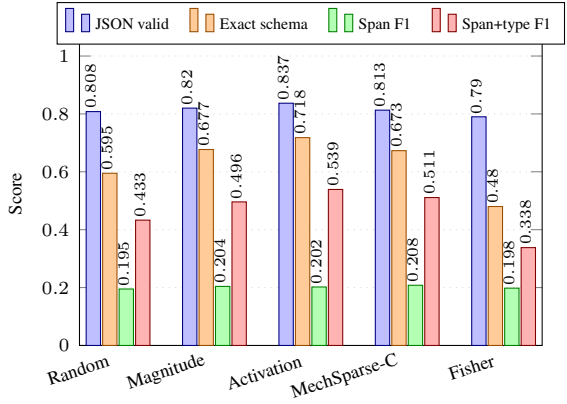
\begin{figure}[t]
\centering
\begin{tikzpicture}
\begin{axis}[
  width=\linewidth, height=5.6cm,
  ybar=0.8pt, bar width=5.4pt,
  symbolic x coords={Random, Magnitude, Activation, MechSparse-C, Fisher},
  xtick=data,
  x tick label style={rotate=18, anchor=east, font=\scriptsize},
  ymin=0, ymax=1.05,
  ytick={0,0.2,0.4,0.6,0.8,1.0},
  ylabel={Score}, ylabel style={font=\scriptsize},
  yticklabel style={font=\scriptsize},
  legend style={font=\tiny, at={(0.5,1.13)}, anchor=north, legend columns=4, /tikz/every even column/.append style={column sep=0.2cm}},
  enlarge x limits=0.10,
  major x tick style=transparent,
  ymajorgrids, grid style={dotted, gray!30},
  nodes near coords,
  nodes near coords style={font=\tiny, rotate=90, anchor=west, /pgf/number format/fixed, /pgf/number format/precision=3, inner sep=1pt},
  every node near coord/.append style={yshift=1pt},
]
\addplot[fill=blue!25,draw=blue!50!black] coordinates {(Random,0.808) (Magnitude,0.820) (Activation,0.837) (MechSparse-C,0.813) (Fisher,0.790)};
\addplot[fill=orange!40,draw=orange!60!black] coordinates {(Random,0.595) (Magnitude,0.677) (Activation,0.718) (MechSparse-C,0.673) (Fisher,0.480)};
\addplot[fill=green!30,draw=green!50!black] coordinates {(Random,0.195) (Magnitude,0.204) (Activation,0.202) (MechSparse-C,0.208) (Fisher,0.198)};
\addplot[fill=red!30,draw=red!50!black] coordinates {(Random,0.433) (Magnitude,0.496) (Activation,0.539) (MechSparse-C,0.511) (Fisher,0.338)};
\legend{JSON valid, Exact schema, Span F1, Span+type F1}
\end{axis}
\end{tikzpicture}
\caption{Selector comparison on Swahili span-JSON IE ($b{=}0.25\%$). Bars show means over 3 seeds (13, 29, 47), $n{=}200$/seed. Per-metric std and the single-seed probe-delta row appear in Appendix~\ref{app:mechsparse_remaining} (Table~\ref{tab:ie_pilot_results}).}
\label{fig:ie_main}
\end{figure}

\paragraph{Headline IE pilot.}
We summarize the Swahili span-JSON IE pilot at budget $0.25\%$ in Figure~\ref{fig:ie_main}, reporting means over three training seeds (13, 29, 47) for the five seed-replicated selectors: random, magnitude, activation-norm, \methodc{}, and gradient/Fisher. We give the per-metric mean$\pm$std and the single-seed probe-delta row in Appendix~\ref{app:mechsparse_remaining} (Table~\ref{tab:ie_pilot_results}). On span+type F1 the selectors separate cleanly: activation-norm is strongest ($0.539{\pm}0.004$), \methodc{} is second with the tightest cross-seed spread ($0.511{\pm}0.003$), magnitude is third ($0.496{\pm}0.057$), random fourth ($0.433{\pm}0.034$), and gradient/Fisher last ($0.338{\pm}0.045$). Paired bootstrap over the pooled $600$ per-example predictions (Table~\ref{tab:ie_bootstrap}) confirms the key comparisons.

\begin{table}[t]
\centering
\scriptsize
\setlength{\tabcolsep}{3pt}
\begin{tabularx}{\linewidth}{@{}L{2.65cm}C{1.05cm}C{2.20cm}C{0.95cm}@{}}
\toprule
\thead{Comparison (S+T F1)} & \thead{Mean $\Delta$\upbetter} & \thead{95\% boot.\ CI} & \thead{$P_+$\upbetter} \\
\midrule
\methodc{} $-$ Random           & $+$0.079 & $[+0.049, +0.107]$ & 1.000 \\
\methodc{} $-$ Gradient/Fisher  & $+$0.174 & $[+0.136, +0.210]$ & 1.000 \\
\methodc{} $-$ Magnitude        & $+$0.015 & $[-0.011, +0.042]$ & 0.873 \\
\methodc{} $-$ Activation-norm  & $-$0.028 & $[-0.048, -0.007]$ & 0.004 \\
Activation $-$ Random           & $+$0.106 & $[+0.078, +0.135]$ & 1.000 \\
\bottomrule
\end{tabularx}
\caption{Paired bootstrap of span+type F1 deltas on the IE pilot, pooled over $600$ predictions across seeds 13, 29, 47 ($2000$ resamples). $P_+=P(\Delta{>}0)$.}
\label{tab:ie_bootstrap}
\end{table}

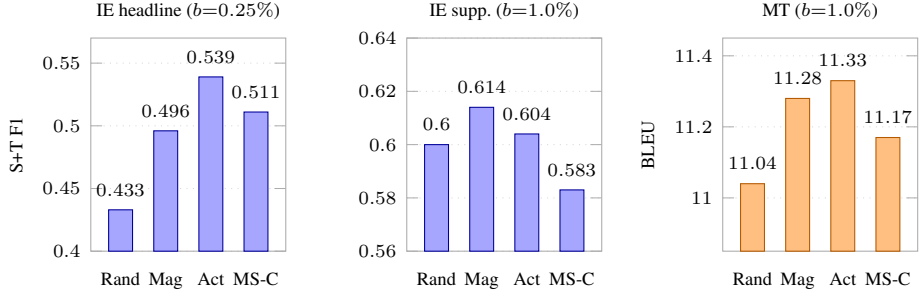
\begin{figure*}[t]
\centering
\begin{tikzpicture}
\begin{groupplot}[
  group style={group size=3 by 1, horizontal sep=1.6cm},
  width=0.26\linewidth, height=4.4cm,
  ybar,
  symbolic x coords={Rand, Mag, Act, MS-C},
  xtick=data,
  x tick label style={font=\scriptsize},
  yticklabel style={font=\scriptsize},
  ylabel style={font=\scriptsize},
  title style={font=\scriptsize, yshift=-0.6ex},
  enlarge x limits=0.22,
  major x tick style=transparent,
  ymajorgrids,
  grid style={dotted, gray!30},
  axis line style={gray!65},
  tick style={gray!65},
  every node near coord/.append style={font=\scriptsize, yshift=1.5pt},
]
\nextgroupplot[ylabel={S+T F1}, title={IE headline ($b{=}0.25\%$)}, ymin=0.40, ymax=0.57, ytick={0.40,0.45,0.50,0.55}, nodes near coords, nodes near coords style={/pgf/number format/fixed, /pgf/number format/precision=3}]
\addplot[bar width=9pt, fill=blue!35,draw=blue!60!black] coordinates {(Rand,0.433) (Mag,0.496) (Act,0.539) (MS-C,0.511)};
\nextgroupplot[title={IE supp.\ ($b{=}1.0\%$)}, ymin=0.56, ymax=0.64, ytick={0.56,0.58,0.60,0.62,0.64}, nodes near coords, nodes near coords style={/pgf/number format/fixed, /pgf/number format/precision=3}]
\addplot[bar width=9pt, fill=blue!35,draw=blue!60!black] coordinates {(Rand,0.600) (Mag,0.614) (Act,0.604) (MS-C,0.583)};
\nextgroupplot[ylabel={BLEU}, title={MT ($b{=}1.0\%$)}, ymin=10.85, ymax=11.45, ytick={11.0,11.2,11.4}, nodes near coords, nodes near coords style={/pgf/number format/fixed, /pgf/number format/precision=2}]
\addplot[bar width=9pt, fill=orange!50,draw=orange!70!black] coordinates {(Rand,11.04) (Mag,11.28) (Act,11.33) (MS-C,11.17)};
\end{groupplot}
\end{tikzpicture}
\caption{Primary-metric selector ordering across the three cells. Supplementary cells support within-cell ordering only (Appendix~\ref{app:mechsparse_remaining}).}
\label{fig:cross_cell}
\end{figure*}

\paragraph{Cross-cell consistency.}
The two supplementary cells reinforce the same task-shaped pattern (Figure~\ref{fig:cross_cell}). At the higher IE budget we see magnitude lead on span+type F1 ($0.614$) with activation-norm close behind ($0.604$); \methodc{} still beats random on JSON validity ($0.830$ vs.\ $0.820$), exact-schema match ($0.805$ vs.\ $0.790$), and span F1 ($0.236$ vs.\ $0.231$), but trails on span+type F1 (Appendix~\ref{app:mechsparse_remaining}, Table~\ref{tab:appendix_ie_medium}). On MT, all four selectors fall within $0.30$ BLEU and paired bootstrap (Table~\ref{tab:mt_bootstrap}) puts every pairwise CI across zero: \methodc{} reaches $11.17$ BLEU, $+0.13$ over random and within $0.16$ of magnitude ($11.28$) and activation-norm ($11.33$). Across cells, we find \methodc{} consistently improves over random on the auxiliary IE metrics, while activation-norm and magnitude remain the strongest cheap selectors whenever schema or broad activation coverage dominates.

\begin{table}[t]
\centering
\scriptsize
\setlength{\tabcolsep}{3pt}
\begin{tabularx}{\linewidth}{@{}L{2.65cm}C{1.05cm}C{2.20cm}C{0.95cm}@{}}
\toprule
\thead{Comparison (BLEU)} & \thead{Mean $\Delta$\upbetter} & \thead{95\% boot.\ CI} & \thead{$P_+$\upbetter} \\
\midrule
\methodc{} $-$ Random           & $+$0.13 & $[-0.59, +0.81]$ & 0.634 \\
\methodc{} $-$ Activation-norm  & $-$0.17 & $[-0.96, +0.59]$ & 0.338 \\
\methodc{} $-$ Magnitude        & $-$0.12 & $[-0.73, +0.51]$ & 0.359 \\
Activation $-$ Random           & $+$0.30 & $[-0.45, +1.07]$ & 0.773 \\
\bottomrule
\end{tabularx}
\caption{Paired bootstrap of MT BLEU deltas at $b{=}1.0\%$ on the recovered $n{=}200$ predictions, single seed ($2000$ resamples). $P_+=P(\Delta{>}0)$.}
\label{tab:mt_bootstrap}
\end{table}

\paragraph{Summary.}
On the headline 3-seed IE cell, \methodc{} is significantly above random ($\Delta{=}{+}0.079$) and gradient/Fisher ($+0.174$), tied with magnitude, and below activation-norm ($-0.028$), while carrying the tightest cross-seed spread ($\pm0.003$). The supplementary cells preserve the broader pattern: cheap magnitude/activation selectors remain hard to beat on primary metrics, while \methodc{} stays above random on auxiliary IE behavior and within the MT noise band. Thus causal selection neither dominates cheap heuristics on schema-heavy IE nor reduces to noise; it gives a diagnostic for \emph{when} mechanistic localization helps sparse PEFT selection (Section~\ref{sec:analysis}).

\section{Analysis}
\label{sec:analysis}

\subsection{Two Objectives, Two Selector Families}

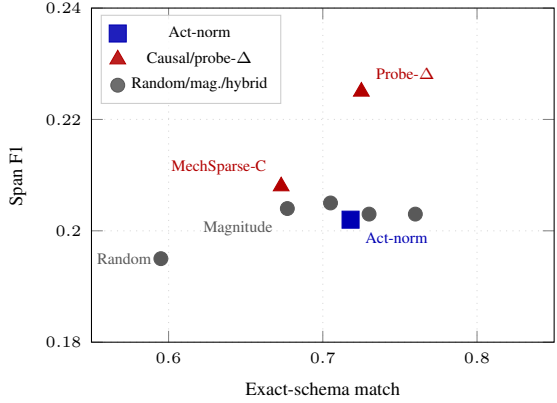
\begin{figure}[t]
\centering
\begin{tikzpicture}
\begin{axis}[
  width=\linewidth, height=6.0cm,
  xlabel={Exact-schema match}, xlabel style={font=\scriptsize},
  ylabel={Span F1}, ylabel style={font=\scriptsize},
  tick label style={font=\tiny},
  xmin=0.55, xmax=0.85,
  ymin=0.180, ymax=0.240,
  legend style={font=\tiny, at={(0.02,0.98)}, anchor=north west, legend columns=1, draw=gray!40, fill=white, fill opacity=0.88, text opacity=1},
  grid=both, grid style={dotted, gray!30},
]
\addplot[only marks, mark=square*, mark size=3.2pt, color=blue!70!black] coordinates {(0.718,0.202)};
\addlegendentry{Act-norm}
\addplot[only marks, mark=triangle*, mark size=3.4pt, color=red!70!black] coordinates {(0.673,0.208) (0.725,0.225)};
\addlegendentry{Causal/probe-$\Delta$}
\addplot[only marks, mark=*, mark size=2.6pt, color=gray!70!black] coordinates {(0.595,0.195) (0.677,0.204) (0.760,0.203) (0.705,0.205) (0.730,0.203)};
\addlegendentry{Random/mag./hybrid}
\node[font=\tiny, text=gray!70!black, anchor=east, xshift=-3pt, fill=white, fill opacity=0.75, text opacity=1, inner sep=0.4pt] at (axis cs:0.595,0.195) {Random};
\node[font=\tiny, text=gray!70!black, anchor=north east,xshift=-5pt,yshift=-4pt, fill=white, fill opacity=0.75, text opacity=1, inner sep=0.4pt] at (axis cs:0.677,0.204) {Magnitude};
\node[font=\tiny, text=blue!70!black, anchor=north west,xshift=5pt,yshift=-4pt, fill=white, fill opacity=0.75, text opacity=1, inner sep=0.4pt] at (axis cs:0.718,0.202) {Act-norm};
\node[font=\tiny, text=red!70!black, anchor=south east,xshift=-5pt,yshift=4pt, fill=white, fill opacity=0.75, text opacity=1, inner sep=0.4pt] at (axis cs:0.673,0.208) {MechSparse-C};
\node[font=\tiny, text=red!70!black, anchor=south west,xshift=5pt,yshift=4pt, fill=white, fill opacity=0.75, text opacity=1, inner sep=0.4pt] at (axis cs:0.725,0.225) {Probe-$\Delta$};
\end{axis}
\end{tikzpicture}
\caption{Schema-vs-span trade-off on IE ($b{=}0.25\%$). Activation-norm, \methodc{}, random, and magnitude are 3-seed means. Probe-$\Delta$ and the three unlabeled gray hybrid/gated points are single-seed. Gradient/Fisher (exact-schema $0.480$) is off the left edge.}
\label{fig:schema_vs_span}
\end{figure}

We read the IE pilot as two objectives bundled into one metric: produce content-correct entities, and emit a rigid JSON schema. Figure~\ref{fig:schema_vs_span} plots the trade-off between exact-schema match ($x$) and span F1 ($y$). We see causal/probe selectors (\methodc{}, probe-delta) occupy the high-span region, while activation-norm anchors the high-schema region. This decomposition explains why the strongest selector depends on which objective the metric rewards.

\paragraph{Schema vs.\ span.}
Schema-heavy generation repeatedly produces braces, keys, list delimiters, type strings, and character offsets. Large-activation components on task prompts tend to carry exactly these always-needed routines, so activation-norm spends its budget on schema-preserving sites. Clean/corrupt patch scores instead reward components whose output shifts when the input entity shifts. We see the split in the numbers: \methodc{} edges activation-norm on span F1 ($0.208{\pm}0.012$ vs.\ $0.202{\pm}0.015$), and the probe-delta point reaches the highest span F1 ($0.225$) and JSON validity ($0.855$). The causal signal is doing what we designed it to do. The primary IE metric, however, also rewards stable type labels carried by activation-heavy components, so activation-norm wins span+type F1. On MT, free-form generation has no rigid schema to preserve, and we see the selectors collapse to under $0.3$ BLEU with every pairwise CI crossing zero (Table~\ref{tab:mt_bootstrap}). We read this as schema-pressure removal, but with one MT seed at BLEU $\sim$11 we cannot rule out a competing floor-effect reading. The IE-vs-MT contrast is therefore suggestive rather than confirmed by our cells. The broader point is that ``important'' is metric-relative: the right selector mirrors the structure of the loss the adapter will be trained against, not an intrinsic property of the network.

\paragraph{Budget effect at IE $b{=}1.0\%$.}
At the higher IE budget we see \methodc{} lead random on JSON validity, exact schema, and span F1, but random pulls ahead on span+type F1 ($0.600$ vs.\ $0.583$). We read the reversal as metric-specific and consistent with budget saturation: at four times the budget, magnitude and activation-norm absorb more schema-preserving sites, leaving \methodc{} a narrower margin while it continues to emphasize content-sensitive sites. The looser supplementary IE prompt amplifies type-string errors, which hurts the span+type composite. Our headline 3-seed result, where \methodc{} beats random by $0.079$ with a fully positive CI, is the regime where site choice has the most leverage.

\subsection{Stability Diagnostic}

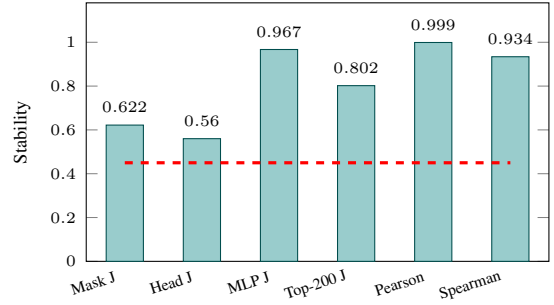
\begin{figure}[t]
\centering
\begin{tikzpicture}
\begin{axis}[
  width=\linewidth, height=5.0cm,
  ybar, bar width=14pt,
  symbolic x coords={Mask J, Head J, MLP J, Top-200 J, Pearson, Spearman},
  xtick=data,
  x tick label style={rotate=18, anchor=east, font=\tiny},
  yticklabel style={font=\tiny},
  ymin=0, ymax=1.18,
  ytick={0,0.2,0.4,0.6,0.8,1.0},
  ylabel={Stability}, ylabel style={font=\scriptsize},
  nodes near coords, nodes near coords style={font=\tiny, yshift=1pt, /pgf/number format/fixed, /pgf/number format/precision=3},
  enlarge x limits=0.10,
  major x tick style=transparent,
]
\addplot[fill=teal!40,draw=teal!60!black] coordinates {(Mask J,0.622) (Head J,0.560) (MLP J,0.967) (Top-200 J,0.802) (Pearson,0.999) (Spearman,0.934)};
\draw[red, dashed, very thick] (axis cs:Mask J,0.45) -- (axis cs:Spearman,0.45);
\end{axis}
\end{tikzpicture}
\caption{Cross-seed stability of single-score \method{} selection on the IE pilot (seed 29 vs.\ base). Dashed line: preregistered $0.45$ floor.}
\label{fig:stability}
\end{figure}

The selector policy only matters if the causal scores themselves are reproducible. We re-ran the single-score \method{} pass on seed 29 and merged it with the base run, then computed six Jaccard/correlation diagnostics on the resulting masks (Figure~\ref{fig:stability}). All six clear the preregistered $0.45$ floor.

We see MLP selection is highly stable (Jaccard $0.967$), and rank order is nearly identical across seeds (Pearson $0.999$). We attribute the residual instability to tie-breaking among similar-scored attention heads near the budget boundary (head Jaccard $0.560$). In practice, aggregate selector behavior -- which layers and which MLPs receive budget -- is reproducible across seeds, while the exact head set near the budget cut-off remains seed-sensitive.

\subsection{Where Do Selectors Disagree?}
\label{sec:selector_disagreement}

Do the selectors merely reorder the same sites, or do they pick genuinely different parts of the network? We compute pairwise mask Jaccard between the four seed-replicated selectors on the headline IE cell, broken out by component type (Table~\ref{tab:mask_overlap}), and the share of selections that fall in the late layers (Table~\ref{tab:late_share}).

\begin{table}[t]
\centering
\scriptsize
\setlength{\tabcolsep}{6pt}
\begin{tabularx}{\linewidth}{@{}L{2.55cm}>{\centering\arraybackslash}X>{\centering\arraybackslash}X>{\centering\arraybackslash}X@{}}
\toprule
\thead{Pair} & \thead{All J\upbetter} & \thead{Head J\upbetter} & \thead{MLP J\upbetter} \\
\midrule
MS-C vs.\ Act.       & 0.260 & 0.160 & 0.778 \\
MS-C vs.\ Mag.       & 0.203 & 0.079 & 0.861 \\
MS-C vs.\ Rand.      & 0.080 & 0.051 & 0.344 \\
Act.\ vs.\ Mag.      & 0.367 & 0.245 & 0.917 \\
\bottomrule
\end{tabularx}
\caption{Pairwise mask Jaccard on IE ($b{=}0.25\%$), overall and restricted to heads/MLPs.}
\label{tab:mask_overlap}
\end{table}

\begin{table}[t]
\centering
\scriptsize
\setlength{\tabcolsep}{4pt}
\begin{tabularx}{\linewidth}{@{}L{2.4cm}C{0.8cm}C{0.8cm}C{0.8cm}C{0.9cm}@{}}
\toprule
\thead{Selector} & \thead{Rand.} & \thead{Mag.} & \thead{Act.} & \thead{MS-C} \\
\midrule
Late-layer share & 37\% & 37\% & 65\% & 46\% \\
\bottomrule
\end{tabularx}
\caption{Share of selected components in late layers (L24--L35) of the 36-layer model.}
\label{tab:late_share}
\end{table}

Both views agree: the selectors target different parts of the model. We find that \methodc{} and activation-norm share roughly $78\%$ of MLP selections but only $\sim$$16\%$ of heads. Activation-norm is late-layer-biased ($65\%$ of its selections fall in L24--35), whereas \methodc{} distributes more uniformly ($46\%$ late). The overlap pattern fits our schema-vs-span reading: the two selectors agree on the high-level MLP sites that route both content and format, but diverge on attention heads near the content-selection boundary, which is exactly where the IE primary metric splits the two objectives.

A second view comes from per-example contingency on the $600$ paired predictions with activation-norm and random (Table~\ref{tab:pred_contingency}): for each example, we check JSON-parseability and the loose schema (top-level dict, \texttt{entities} list with \texttt{span} and \texttt{type}).

\begin{table}[t]
\centering
\scriptsize
\setlength{\tabcolsep}{3pt}
\begin{tabularx}{\linewidth}{@{}L{3.20cm}C{0.65cm}C{0.85cm}C{0.85cm}C{0.85cm}@{}}
\toprule
\thead{Comparison} & \thead{Both\upbetter} & \thead{A\upbetter} & \thead{B\upbetter} & \thead{None\downbetter} \\
\midrule
JSON valid \\
\quad MS-C vs.\ Act.   & 468 & 20 & 34 & 78 \\
\quad MS-C vs.\ Rand.  & 445 & 43 & 40 & 72 \\
\midrule
Loose schema \\
\quad MS-C vs.\ Act.   & 360 & \textbf{43} & 24 & 173 \\
\quad MS-C vs.\ Rand.  & 266 & \textbf{137} & 34 & 163 \\
\bottomrule
\end{tabularx}
\caption{Per-example prediction contingency, pooled over seeds 13/29/47 ($600$ examples) for paired-prediction selectors. A-only / B-only count examples where exactly one selector's output is JSON-parseable (top) or matches the loose schema (bottom).}
\label{tab:pred_contingency}
\end{table}

On the loose schema, we find \methodc{} significantly beats activation-norm (McNemar $p{=}0.027$). On the stricter exact-schema metric (Figure~\ref{fig:ie_main}), activation-norm wins. We trace the gap to the offset/type-string columns, not the list structure -- consistent with activation-norm picking late-layer components that carry stable type-label routines.

\subsection{Selector Decision Rule}

We distill the findings into preliminary guidance whose branches carry uneven evidence. When the task has rigid output structure and the bottleneck components are always-on, we recommend \textbf{activation-norm}: it is the cheapest selector and wins schema-heavy IE under multi-seed CIs. For free-form generation, the choice barely matters in our MT cell ($<0.3$ BLEU spread, CIs cross zero), a result also consistent with a floor effect at BLEU $\sim$11. When span localization or content sensitivity is primary, \textbf{probe-delta} reaches the best JSON validity ($0.855$) and span F1 ($0.225$) in our headline cell, but it is a single-seed point whose lead falls within the cross-seed std of activation-norm and \methodc{} on span F1 ($\pm 0.012$--$0.015$). We therefore offer it as a \emph{hypothesis} for content-dominated cells rather than a default. We reserve full \methodc{} for tasks where the application justifies the patching compute and local joint subsets are expected to matter.

\section{Related Work}

\paragraph{Mechanistic localization.}
Attention-head analysis and causal intervention show that transformer components can specialize in syntactic, positional, factual, and task-relevant behavior \citep{clark2019bert,jawahar2019bertology,voita2019analyzing,michel2019sixteen,geva2021transformer}. Causal mediation, causal tracing, and activation patching provide intervention-based evidence beyond correlational attribution \citep{vig2020investigating,meng2022locating,conmy2023automated,zhang2023patching}. We use these tools as a \emph{selection signal} for adaptation rather than as an endpoint: the intervention score decides where we place the sparse adapters.

\paragraph{Pitfalls of patching.}
\citet{zhang2023patching} and \citet{heimersheim2024patching} document that single-component patch scores are sensitive to (i) the choice of corruption operator, (ii) the normalization denominator when the base model is already correct on the probe, and (iii) interactions with neighboring components (the ``hydra effect''). We address these directly with a base-model competence filter on probes (\S\ref{sec:method}), an $\epsilon$-stabilized denominator, and the bounded joint-subset variant \methodc{} that scores small within-layer subsets. Our IE result adds a separate caveat: a causal selector can improve over random while still losing to an activation heuristic when the downstream task rewards schema stability.

\paragraph{Sparse and masked PEFT.}
Adapters, prefix/prompt tuning, BitFit, IA3, LoRA, QLoRA, AdaLoRA, SoRA, and Fourier-PEFT reduce trainable parameters but choose sites by architecture or by optimization heuristics \citep{houlsby2019parameter,li2021prefix,lester2021power,zaken2022bitfit,liu2022ia3,hu2022lora,dettmers2023qlora,zhang2023adalora,ding2023sora,gao2024fourier}. Closest in spirit are optimization-derived selectors. AdaLoRA reallocates rank across layers using the singular-value importance of the running adapter \citep{zhang2023adalora}, SoRA uses gating that drives less-useful rank dimensions to zero \citep{ding2023sora}, and Fisher- or Taylor-importance methods score each component by an estimate of its loss gradient. These methods all read importance from the \emph{training objective on the target task}. We instead read importance from a \emph{causal} probe defined before training -- which components, if their clean-run activations were patched into a corrupted run, would restore the clean target -- and use that score as a selection signal. The two families are complementary: a causal selector identifies sites that matter for a specific behavior, while a Fisher/AdaLoRA selector identifies sites whose updates reduce task loss most quickly. We compare against gradient/Fisher directly on the headline IE cell (Section~\ref{sec:results}); AdaLoRA and Taylor-importance comparisons remain future work (Section~\ref{sec:limitations}). \citet{gao2025weightsparse} show that weight-sparse transformers yield more interpretable circuits, but do not test masks as a PEFT signal. We ask the complementary question: can \emph{causal} patching scores choose useful PEFT sites under a matched budget?

\paragraph{Training-time vs.\ inference-time use of causal scores.}
Activation patching and causal mediation have been used primarily at \emph{inference time}: to identify circuits responsible for a behavior \citep{meng2022locating,conmy2023automated}, to steer outputs by editing activations of specific heads, or to ablate harmful behavior without retraining. We port these signals to \emph{training time} and use them as a site-selection rule for sparse adapters. The two regimes ask different questions of the same score: inference-time control wants components whose intervention changes a behavior immediately, while training-time selection wants components whose adapter updates will improve a behavior after gradient descent. Our results show why the gap matters: causal masks identify content-sensitive sites, but schema-heavy losses also reward the always-on formatting routines that patching does not flag.

\paragraph{Multilingual and low-resource adaptation.}
Multilingual models expose shared and language-specific structure \citep{aharoni2019massively,fan2021beyond,liu2020mbart,tang2020multilingual,conneau2020xlmr,xue2021mt5,muller2021when}. Low-resource African MT and NER benchmarks anchor falsifiable transfer questions \citep{goyal2022flores,nllb2022,adelani2022masakhaner,nekoto2020participatory}. We measure on Swahili MasakhaNER span-JSON at two budgets and on English$\to$Swahili MT at one budget. The multilingual framing motivates the choice of cells, but the cross-task finding (selector quality is task-shaped) is the headline claim, not a Swahili-specific result.

\paragraph{Cost-aware empirical evaluation.}
Because the claim is about transfer under constrained resources, quality metrics alone are insufficient. Our pilot reports trainable parameter count, component count, schema validity, exact-schema match, span F1, and span+type F1, and ties each numeric value to released result artifacts. Future MT runs should follow MT practice in reporting SacreBLEU \citep{post2018sacrebleu}, chrF \citep{popovic2015chrf}, COMET \citep{rei2020comet}, XCOMET-XL \citep{guerreiro2024xcomet}, and matched significance tests \citep{koehn2004statistical,dror2018hitchhiker,bergkirkpatrick2012empirical}.

\section{Conclusion}

\method{} reports a diagnostic: across three Ministral-8B/NF4 cells, mechanistic localization does not beat cheap heuristics on the primary metric. On the headline 3-seed IE cell, \methodc{} significantly beats random and gradient/Fisher, ties magnitude, has the tightest cross-seed spread ($\pm 0.003$), but trails activation-norm by $0.028$ span+type F1. The schema-vs-span decomposition (Section~\ref{sec:analysis}) explains the IE pattern and yields a preliminary rule: prefer activation-norm under rigid output structure, and treat causal/probe selectors as a hypothesis for content-dominated cells not yet demonstrated here. We release the masks, selector scores, predictions, and per-seed evaluation files so that the comparisons can be rerun directly under our protocol. The most informative follow-ups would identify a cell where a causal selector wins a primary metric with a positive multi-seed CI, measure localization per task, and add adaptive-rank baselines (AdaLoRA, Taylor). Selectors track the loss, not the network.

\clearpage
\section*{Limitations}
\label{sec:limitations}

\paragraph{No positive witness for the causal branch of the rule.}
Across all three cells we measure, the proposed causal selectors do not win the primary metric. The decision rule's ``use causal/probe selectors when content sensitivity is the bottleneck'' branch is therefore a hypothesis, not a demonstrated finding. Confirming it requires at least one cell where a causal selector wins a primary metric with a fully positive multi-seed CI.

\paragraph{Mechanism-localization premise is unmeasured.}
\method{} assumes the task mechanism is sparse enough to be worth targeting. We do not directly measure this on our cells (e.g.\ Gini/entropy of patch scores per task). If the IE or MT mechanism is not sparsely localized, the underlying framing is ill-posed and could itself explain why a causal mask does not beat cheap heuristics.

\paragraph{Pooled-prediction CIs vs.\ seed-level variance.}
Headline paired bootstrap pools the $600$ per-example predictions across the three seeds and resamples examples. With only three seeds, an explicit seed-level test is not powered; the example-level CIs can therefore understate between-seed uncertainty. We mitigate this by reporting per-selector cross-seed stds (Appendix~\ref{app:mechsparse_remaining}, Table~\ref{tab:ie_pilot_results}) alongside the CIs and by foregrounding the tighter-spread selectors in the prose.

\paragraph{Score-to-placement granularity.}
Our patch scores are per-query-head, but the base model's grouped-query attention shares $Q$/$O$ projections across all heads and $K$/$V$ across each $4$-head group, so our projection-level LoRA is coarser than the score. A per-head row-slice LoRA would let us test whether finer placement increases the value of the same causal ranking.

\paragraph{Bounded joint subsets and selector hyperparameters.}
We run \methodc{} with subsets of size $\le 2$, $\le 8$ per layer, and $\lambda{=}0.4$, which we consider a practical budget for local joint credit. Cross-layer subsets, larger subsets, and sweeps over the corruption operator, $\epsilon$, probe count, per-layer cap ($40\%$), or type floor ($20\%$) would let us map the selector's sensitivity.

\paragraph{Seed-replicated evidence, MT power, and rank-allocation baselines.}
Three end-to-end seeds cover the headline IE cell. Supplementary IE and MT are single-seed; for MT we additionally rely on BLEU/chrF without COMET/XCOMET, and selector convergence at BLEU $\sim$11 is consistent with both schema-pressure removal and a floor effect. Probe-delta is also single-seed, so its span-F1 and JSON-validity lead falls within the cross-seed stds of the alternatives. AdaLoRA-style adaptive-rank and Taylor-importance selectors are complementary rank-allocation baselines we leave to future work.

\paragraph{Selector-asymmetric task exposure.}
Activation-norm, \methodc{}, and probe-delta all see $64$ task prompts (no test data), while random sees nothing. Activation-norm benefits from \emph{seeing the task format at all}: high-activation components on schema-rich prompts coincide with format-preserving sites. A task-agnostic activation-norm variant would isolate this effect.

\paragraph{Joint-subset stability and probe design.}
The seed-29 diagnostic covers single-score \method{} stability only. The 3-seed metric variance ($0.511{\pm}0.003$) bounds the downstream effect of any residual instability in the joint-subset masks. Probe construction determines what the score sees, and the IE corruptions used here target the content movement evaluated by span F1.

\paragraph{Scope of cells.}
We measure three task cells with one model. The supplementary IE cell uses a looser prompt, both supplementary cells predate GQA K/V dedup, and they support within-cell ordering only. Additional model families, languages, and at least one non-schema-heavy structured task (morphology, code-switched IE, light-schema summarization) would test the decision rule across a wider design space.

\section*{Ethics and Broader Impact}

Resource-efficient adaptation can lower barriers for low-resource NLP, but poor structured extraction can harm users if deployed without review. This pilot is a research diagnostic, not a deployable NER system. MasakhaNER examples carry public named-entity annotations, but generated structured outputs should not be used to infer real-world facts about people. Released masks and summaries include model licenses, intended use, language coverage, and a warning that mechanistic labels are approximate. No new human annotation study is reported. All empirical claims derive from local evaluation artifacts.

\clearpage
\appendix
\begingroup
\emergencystretch=3em
\section{Reproducibility and Hyperparameters}
\label{app:hyperparams}

\paragraph{Appendix guide.}
We use this appendix as the audit trail for the numerical claims. Appendix~\ref{app:hyperparams} lists the released artifacts and primary-run settings. Appendix~\ref{app:probe_schema_prompt} defines the IE target, prompt, and corruption protocol. Appendix~\ref{app:mechsparse_remaining} reports diagnostic runs used to interpret selector behavior. The main text states which cells support headline claims and which cells support within-cell ordering.

\paragraph{Released artifacts.}
The release separates primary evidence from supplementary diagnostics. For the current-pipeline IE cell, we release adapter weights, masks, selector scores, $200$-example test predictions, and per-method \texttt{eval.json} files for all five seed-replicated selectors across seeds $13, 29, 47$. The $3$-seed aggregate JSON backs Table~\ref{tab:ie_pilot_results}.

The supplementary MT cell includes recovered $200$-example predictions and per-selector \texttt{eval.json} files from the earlier pipeline. The supplementary IE cell's \texttt{eval.json} files are not in the release. Its values are preserved in two released synthesized reports: a timestamped status report and a baseline audit.

Every numeric value in the main figures and result tables traces to these files. The bootstrap tables use the released predictions with $2000$ paired resamples. Compact JSON summaries index each paper claim to its source artifact.

\paragraph{Configuration.}
The primary IE run uses the following settings from the released run scenario:

\begin{center}
\centering
\footnotesize
\setlength{\tabcolsep}{3pt}
\begin{tabularx}{\linewidth}{@{}L{1.45cm}L{5.0cm}@{}}
\toprule
\thead{Field} & \thead{Setting} \\
\midrule
Model & Ministral-8B Instruct with NF4 quantization and bf16 compute. \\
Task & Swahili span-JSON IE from MasakhaNER (\texttt{swa}), train limit $1024$, eval limit $200$. \\
Probe filter & Clean probability $\ge 0.55$, corrupt probability $\le 0.45$. \\
Patching & $64$ probes, $\epsilon{=}0.05$, $100$ bootstrap samples, joint top fraction $0.20$, max subset size $2$, mix $\lambda{=}0.4$. \\
Budget & $b{=}0.25\%$, head rank $4$, MLP rank $8$, per-layer cap $0.40$, type floor $0.20$. \\
Training & One epoch, learning rate $2{\times}10^{-4}$, gradient accumulation $16$, max length $768$, bf16. \\
Evaluation & Test split, max $96$ new tokens, generation batch size $8$. \\
\bottomrule
\end{tabularx}
\end{center}

\paragraph{Evaluation files.}
At evaluation time, each selector produces an \texttt{eval.json} per cell. The core metrics in Table~\ref{tab:ie_pilot_results} and the exploratory metrics in Appendix~\ref{app:mechsparse_remaining} are exactly the values in those files and the released summary reports. We retain the exploratory hybrid/gated methods because they document completed runs and explain the gray cluster in Figure~\ref{fig:schema_vs_span}. They are diagnostics rather than tuned baselines.

\paragraph{Large artifacts.}
Adapter weights, raw predictions, processed datasets, and logs are released as a separate archive (roughly $245$ MB). The paper repository carries only the compact numeric summaries needed to audit the manuscript claims.

\section{Probe, Schema, and Prompt Protocol}
\label{app:probe_schema_prompt}

This appendix defines the structured-generation target and the clean/corrupt probes used by the patching selector. IE evaluation combines span recovery with rigid JSON-schema preservation, so the schema-vs-span analysis in Section~\ref{sec:analysis} separates the two behaviors explicitly.

\paragraph{Span-JSON IE schema.}
Each MasakhaNER example in the pilot is converted to a span-JSON target. The required output follows this grammar:
\begin{verbatim}
{
  "entities": [
    {"span": "<surface>",
     "type": "PER|ORG|LOC|DATE",
     "start": <char index>,
     "end":   <char index>}
  ]
}
\end{verbatim}
The model receives the input sentence and must output only the \texttt{entities} list. JSON validity is judged by a strict parser. Exact-schema match additionally requires the expected keys and value types. Span F1 rewards correct character spans, while span+type F1 requires both the span and the entity type to match. Thus a prediction can improve span localization while still losing the primary metric if it mislabels entity types or breaks the schema.

\paragraph{Prompt format.}
The pilot uses chat-style prompts with the system message:
\begin{quote}
\small
\raggedright
You are a professional translator and information extraction assistant.
\end{quote}
The user prompt asks for strict JSON only, names the allowed entity types, specifies zero-based character offsets with exclusive end positions, and supplies the source text. The generation output is parsed directly, so markdown and explanatory text are disallowed. This strict parsing is intentional: it makes schema preservation an evaluated behavior rather than a cosmetic formatting preference.

\paragraph{Corruptions.}
IE corruptions create the contrast needed for patching by replacing entity spans or altering entity types, which changes the expected JSON target. The clean/corrupt filter keeps only probes where the base model assigns sufficient probability to the clean target and the corrupted prompt lowers that clean-target probability. The filter ensures that patch recovery measures a real contrast in the model, not examples where the base model failed the clean case.

\paragraph{Reporting protocol.}
Numeric claims in this revision come from the artifacts named in Appendix~\ref{app:hyperparams}. The paper-facing claim ledger records each manuscript claim, its status, and its source artifact. The original run ledger remains in the ignored EC2 download folder and contains one row per raw metric.

\FloatBarrier
\section{Remaining MechSparse Results}
\label{app:mechsparse_remaining}

The main paper reports the headline IE pilot, the cross-cell selector ordering (Figure~\ref{fig:cross_cell}), and the seed-29 stability diagnostic. This appendix keeps the supporting evidence visible. Table~\ref{tab:ie_pilot_results} gives the per-metric mean$\pm$std backing Figure~\ref{fig:ie_main}. Table~\ref{tab:appendix_ie_variants} gives exploratory primary-cell variants that explain the gray points in Figure~\ref{fig:schema_vs_span}. These variants are diagnostics rather than tuned baselines. Tables~\ref{tab:appendix_mt_medium}--\ref{tab:appendix_ie_medium} give the two supplementary cells used for within-cell selector ordering.

\begin{table}[!htbp]
\centering
\scriptsize
\setlength{\tabcolsep}{2pt}
\begin{tabularx}{\linewidth}{@{}L{1.60cm}C{1.18cm}C{1.18cm}C{1.18cm}C{1.18cm}@{}}
\toprule
\thead{Method} & \thead{JSON\upbetter} & \thead{Exact\upbetter} & \thead{Span F1\upbetter} & \thead{S+T F1\upbetter} \\
\midrule
Random           & 0.808\tiny{$\pm$.014} & 0.595\tiny{$\pm$.038} & 0.195\tiny{$\pm$.007} & 0.433\tiny{$\pm$.034} \\
Magnitude        & 0.820\tiny{$\pm$.020} & 0.677\tiny{$\pm$.103} & 0.204\tiny{$\pm$.019} & 0.496\tiny{$\pm$.057} \\
Act.-norm        & \textbf{0.837}\tiny{$\pm$.015} & \textbf{0.718}\tiny{$\pm$.050} & 0.202\tiny{$\pm$.015} & \textbf{0.539}\tiny{$\pm$.004} \\
\methodc{}       & 0.813\tiny{$\pm$.041} & 0.673\tiny{$\pm$.037} & \textbf{0.208}\tiny{$\pm$.012} & 0.511\tiny{$\pm$.003} \\
Grad./Fisher     & 0.790\tiny{$\pm$.018} & 0.480\tiny{$\pm$.123} & 0.198\tiny{$\pm$.008} & 0.338\tiny{$\pm$.045} \\
Probe-$\Delta$   & \textbf{0.855} & 0.725 & \textbf{0.225} & 0.486 \\
\bottomrule
\end{tabularx}
\caption{Headline IE pilot at $b{=}0.25\%$ on Swahili MasakhaNER span-JSON. The first five rows report mean$\pm$std over $3$ seeds (13, 29, 47), with $200$ test examples per seed. Probe-$\Delta$ is a single-seed diagnostic. All methods are matched to $\sim$15M trainable parameters.}
\label{tab:ie_pilot_results}
\end{table}

\begin{table}[!htbp]
\centering
\scriptsize
\setlength{\tabcolsep}{2pt}
\begin{tabularx}{\linewidth}{@{}L{1.55cm}C{0.58cm}C{0.58cm}C{0.58cm}C{0.68cm}C{0.55cm}C{0.82cm}@{}}
\toprule
\thead{Hybrid/gated selector} & \thead{J.} & \thead{Ex.} & \thead{Sp.} & \thead{S+T} & \thead{C.} & \thead{Par.} \\
\midrule
Gated pos. activation & 0.800 & 0.730 & 0.203 & 0.490 & 224 & 15.01M \\
Hybrid $\alpha{=}0.25$ & 0.815 & 0.705 & 0.205 & 0.491 & 222 & 14.97M \\
Hybrid $\alpha{=}0.50$ & 0.815 & 0.760 & 0.203 & 0.491 & 221 & 14.97M \\
\bottomrule
\end{tabularx}
\caption{Exploratory hybrid/gated IE selectors at $b{=}0.25\%$ from the primary cell. These rows correspond to the gray points in Figure~\ref{fig:schema_vs_span}. The positive-activation gate remains below activation-norm and MechSparse-C on the primary metric; the hybrid settings improve some auxiliary metrics but do not close the span+type gap.}
\label{tab:appendix_ie_variants}
\end{table}

\subsection*{Supplementary-cell diagnostic}
\label{app:medium_run_diagnostic}

The cross-cell figure in the main paper (Figure~\ref{fig:cross_cell}) uses two supplementary cells from an earlier pipeline iteration. That run's per-method \texttt{eval.json} files are not in the released bundle, but a final status report and baseline audit record the completed MT BLEU/chrF and the IE table at budget $1.0\%$. The supplementary IE cell used a looser prompt, and both supplementary cells predated GQA K/V mask deduplication. The audit applied the canonical IE evaluator to the earlier IE predictions, while MT BLEU/chrF was unchanged. We therefore compare selectors \emph{within} each cell, not absolute values across pipeline iterations.

The two supplementary cells play a focused role in the argument: they check whether the qualitative selector ordering is brittle. The answer is consistent with the main text. \methodc{} is usually better than random, magnitude and activation-norm remain strong cheap selectors, and MT shows little statistically reliable separation among selectors.

\begin{table}[!htbp]
\centering
\footnotesize
\begin{tabularx}{\linewidth}{@{}L{2.3cm}C{1.2cm}C{1.2cm}@{}}
\toprule
\thead{Method (MT, $b{=}1.0\%$)} & \thead{BLEU\upbetter} & \thead{chrF\upbetter} \\
\midrule
Random & 11.04 & 39.32 \\
Magnitude & \textbf{11.28} & \textbf{39.87} \\
Activation-norm & \textbf{11.33} & 39.23 \\
\methodc{} & 11.17 & 39.32 \\
\bottomrule
\end{tabularx}
\caption{MT panel values for Figure~\ref{fig:cross_cell}: Swahili MT BLEU/chrF at $b{=}1.0\%$ from the earlier run. MechSparse-C beats random on BLEU ($+0.13$) but trails magnitude and activation-norm. This cell supports within-cell ordering only.}
\label{tab:appendix_mt_medium}
\end{table}

\begin{table}[!htbp]
\centering
\scriptsize
\setlength{\tabcolsep}{2pt}
\begin{tabularx}{\linewidth}{@{}L{1.75cm}C{0.72cm}C{0.72cm}C{0.82cm}C{0.82cm}@{}}
\toprule
\thead{Method (IE, $b{=}1.0\%$)} & \thead{JSON\rlap{\upbetter}} & \thead{Exact\rlap{\upbetter}} & \thead{Span\rlap{\upbetter}} & \thead{S+T\rlap{\upbetter}} \\
\midrule
Random & 0.820 & 0.790 & 0.231 & 0.600 \\
Magnitude & 0.840 & \textbf{0.815} & \textbf{0.250} & \textbf{0.614} \\
Activation-norm & \textbf{0.850} & \textbf{0.815} & 0.242 & 0.604 \\
\methodc{} & 0.830 & 0.805 & 0.236 & 0.583 \\
\bottomrule
\end{tabularx}
\caption{Supplementary IE values for the middle panel of Figure~\ref{fig:cross_cell} at $b{=}1.0\%$. The audit applied the canonical IE evaluator, but training used the earlier looser prompt and pre-dedup GQA mask accounting. This cell supports within-cell ordering only.}
\label{tab:appendix_ie_medium}
\end{table}

\endgroup

\end{document}